\documentclass[11pt,a4paper]{article}
\usepackage[hyperref]{acl2020}
\usepackage{times}
\usepackage{latexsym}

\usepackage{microtype}

\aclfinalcopy 

\usepackage{multirow}

\usepackage{graphicx}
\usepackage{tikz}
\usetikzlibrary{arrows, automata, cd, babel}
\usepackage[export]{adjustbox}
\usepackage{subcaption}

\usepackage{amsmath}
\usepackage{amsfonts}
\usepackage{amssymb}
\usepackage{amsthm}
\allowdisplaybreaks

\usepackage{gb4e}
\usepackage{tipa}
\noautomath

\theoremstyle{definition}

\newtheorem{example}[exx]{Example}

\DeclareMathOperator{\argmax}{argmax}

\usepackage{natbib}

\title{Attribution Analysis of Grammatical Dependencies in LSTMs}

\author{Yiding Hao \\
  Yale University\\
  New Haven, CT, USA\\
  \texttt{yiding.hao@yale.edu}}

\date{}

\begin{document}
\maketitle
\begin{abstract}
	LSTM language models have been shown to capture syntax-sensitive grammatical dependencies such as subject--verb agreement with a high degree of accuracy (\citealp{linzenAssessingAbilityLSTMs2016}, \textit{inter alia}). However, questions remain regarding whether they do so using spurious correlations, or whether they are truly able to match verbs with their subjects. This paper argues for the latter hypothesis. Using \textit{layer-wise relevance propagation} \citep{bachPixelWiseExplanationsNonLinear2015}, a technique that quantifies the contribution of input features to model behavior, we show that LSTM performance on number agreement is directly correlated with the model's ability to distinguish subjects from other nouns. Our results suggest that LSTM language models are able to infer robust representations of syntactic dependencies.
\end{abstract}

\section{Introduction}

A major area of research in interpretable NLP concerns the question of whether black-box models are capable of inferring complex grammatical dependencies from data. \citet{linzenAssessingAbilityLSTMs2016} approached this question by constructing a diagnostic task based on subject--verb agreement whose solution requires knowledge of natural language syntax. Since this seminal work, the basic methodology has been extended in several ways. These include hand-crafting testing sets to control for linguistic features \citep{marvinTargetedSyntacticEvaluation2018,wilcoxWhatRNNLanguage2018,warstadtInvestigatingBERTKnowledge2019}; extracting information from network layers using diagnostic classification \citep{giulianelliHoodUsingDiagnostic2018,linOpenSesameGetting2019}; and detecting representations of syntactic structure through unsupervised parsing \citep{merrillFindingHierarchicalStructure2019}. 

Methodologies based on task and testing set design demonstrate that models exhibit behavior consistent with that of a fully interpretable model, and methodologies based on extracting representations demonstrate that model weights contain enough information to represent some aspect of natural language grammar. Supplementing these approaches, we propose the use of \textit{attribution analysis}, a methodology that enables us to directly determine the reasoning by which a model arrives at a certain decision. In attribution analysis, each input to a model is assigned a score measuring the importance of that input in determining the model's output. Interpretable models should assign high scores to input features that are relevant for computing the output, while those that assign high scores to irrelevant features are likely to be ``Clever Hans predictors'' \citep{lapuschkinUnmaskingCleverHans2019}---models that primarily rely on spurious correlations to optimize their training objectives.

This paper adopts \textit{layer-wise relevance propagation} (LRP, \citealp{bachPixelWiseExplanationsNonLinear2015}), an attribution method that distributes logit scores among model inputs, and applies it to \citeauthor{linzenAssessingAbilityLSTMs2016}'s (\citeyear{linzenAssessingAbilityLSTMs2016}) subject--verb agreement task. Using \citeauthor{marvinTargetedSyntacticEvaluation2018}'s (\citeyear{marvinTargetedSyntacticEvaluation2018}) \textit{Targeted Syntactic Evaluation} (TSE), a partitioned testing set that controls for syntactic structure, we show that the performance of LSTM language models on subject--verb agreement is directly correlated with the degree to which subjects are assigned higher scores than other words, while failure to exhibit this phenomenon results in degraded performance. These results show that our model enforces agreement by matching verbs with their subjects, and not by relying on idiosyncratic statistical properties of the training data.

This paper is structured as follows. Sections \ref{sec:lrp} and \ref{sec:relatedwork} review LRP and the relevant literature on attribution analysis. Section \ref{sec:procedure} describes our experimental procedure, and Section \ref{sec:results} presents our results. These results are discussed in Section \ref{sec:discussion}, and Section \ref{sec:conclusion} concludes.

\section{Layer-Wise Relevance Propagation}
\label{sec:lrp}

LRP assigns to each input of a network a \textit{relevance score} representing its contribution to the model output. To understand what this means, let us consider an illustrative example.

\begin{example}
	\label{ex:alice}
	Suppose Alice holds two part-time positions. She works for $x$ hours in position 1 at the rate of \$$p/\text{hour}$ and $y$ hours in position 2 at the rate of \$$q/\text{hour}$. Alice's total income is given by the function
	\[
	f(x, y) = t(px + qy),
	\]
	where $t$ is a non-linear function mapping Alice's gross income to her post-tax income. 
	
	LRP asks the following question: how much of Alice's income comes from position 1 and how much comes from position 2? It is clear that \$$px$ of Alice's \textit{pre-tax} income comes from position 1 and \$$qy$ comes from position 2. Intuitively, we may attribute Alice's \textit{total} income to the two positions in the same proportions. Thus, the amount of money Alice has earned from position 1 is
	\[
	R(x) = f(x, y) \cdot \frac{px}{px + qy},
	\]
	and the amount earned from position 2 is
	\[
	R(y) = f(x, y) \cdot \frac{qy}{px + qy}.
	\]
\end{example}

We can apply this reasoning to LSTMs using the following derivation, due to \citet{arrasExplainingRecurrentNeural2017}. Consider an LSTM classifier that takes inputs $\mathbf{x}_1, \allowbreak \mathbf{x}_2, \allowbreak \dots, \allowbreak \mathbf{x}_t$ and passes the final hidden state $\mathbf{h}_t$ through a linear layer, producing a vector $\mathbf{y}$ of logit scores. We initialize the relevance $R(\mathbf{y})$ of the output layer by 
\[
R(\mathbf{y}) = \argmax(\mathbf{y}).
\]
We seek to determine the contribution of each input $\mathbf{x}_i$ to the logit score $\mathbf{1}^\top R(\mathbf{y})$ of the predicted class. We do this by propagating the relevance value backwards through the network, applying the reasoning of Example \ref{ex:alice} repeatedly.

To begin, we propagate $R(\mathbf{y})$ to the final hidden state $\mathbf{h}_t$ and the linear layer bias term $\mathbf{b}$. Following the the reasoning of Example \ref{ex:alice}, the relevance of each term is determined by the proportion of $\mathbf{y}$ comprised by that term.\footnote{In practice, the denominator also contains a stabilizing term, cf. \citet{arrasExplainingRecurrentNeural2017,arrasEvaluatingRecurrentNeural2019}.}
\[
R(\mathbf{h}_t) = R(\mathbf{y}) \odot \frac{\mathbf{W}\mathbf{h}_t}{\mathbf{y}} = R(\mathbf{y}) \odot \frac{\mathbf{W}\mathbf{h}_t}{\mathbf{W}\mathbf{h}_t + \mathbf{b}}
\]
Recall that $\mathbf{h}_t$ is given by
\begin{align}
	\mathbf{h}_t &= \mathbf{o}_t \odot \tanh(\mathbf{c}_t) \nonumber \\
	&= \mathbf{o}_t \odot \tanh(\mathbf{f}_t \odot \mathbf{c}_{t - 1} + \mathbf{i}_t \odot \mathbf{g}_t), \label{eqn:ctrelevance}
\end{align}
where $\mathbf{c}_t$ is the \textit{cell state} and $\mathbf{f}_t$, $\mathbf{i}_t$, and $\mathbf{o}_t$ are the \textit{forget gate}, \textit{input gate}, and \textit{output gate}, respectively. Following \citet{arrasExplainingRecurrentNeural2017}, we treat the output gate $\mathbf{o}_t$ as a unary operator $\mathbf{v} \mapsto \mathbf{o}_t \odot \mathbf{v}$, so that (\ref{eqn:ctrelevance}) may be viewed as a linear mapping with activation $\mathbf{o}_t \odot \tanh(\cdot)$. Example \ref{ex:alice} then gives us
\[
R(\mathbf{i}_t \odot \mathbf{g}_t) = R(\mathbf{h}_t) \odot \frac{\mathbf{i}_t \odot \mathbf{g}_t}{\mathbf{f}_t \odot \mathbf{c}_{t - 1} + \mathbf{i}_t \odot \mathbf{g}_t}.
\]
Next, we rewrite $\mathbf{i}_t \odot \mathbf{g}_t$ as a linear mapping with activation $\mathbf{i}_t \odot \tanh(\cdot)$:
\[
\mathbf{i}_t \odot \mathbf{g}_t = \mathbf{i}_t \odot \tanh(\mathbf{W}_{g, x}\mathbf{x}_t + \mathbf{W}_{g, h}\mathbf{h}_{t - 1} + \mathbf{b}_g).
\]
By Example \ref{ex:alice},
\[
R(\mathbf{x}_t) = R(\mathbf{i}_t \odot \mathbf{g}_t) \odot \frac{\mathbf{W}_{g, x}\mathbf{x}_t}{\mathbf{W}_{g, x}\mathbf{x}_t + \mathbf{W}_{g, h}\mathbf{h}_{t - 1} + \mathbf{b}_g}
\]
and
\begin{align*}
&\mathrel{\phantom{=}} R(\mathbf{h}_{t - 1}) \\
&= R(\mathbf{i}_t \odot \mathbf{g}_t) \odot \frac{\mathbf{W}_{g, h}\mathbf{h}_{t - 1}}{\mathbf{W}_{g, x}\mathbf{x}_t + \mathbf{W}_{g, h}\mathbf{h}_{t - 1} + \mathbf{b}_g}.
\end{align*}
This completes the backwards relevance propagation for one time step of the LSTM. To compute the relevance propagation for the next time step, we notice from (\ref{eqn:ctrelevance}) that both $\mathbf{h}_t$ and $\mathbf{h}_{t - 1}$ propagate relevance to $\mathbf{c}_{t - 1}$. To account for this, we decompose (\ref{eqn:ctrelevance}) into two separate equations:
\begin{align}
	\mathbf{h}_{t - 1} &= \mathbf{o}_{t - 1} \odot \tanh(\mathbf{c}_{t - 1}) \label{eqn:ctrelevance2} \\
	\mathbf{c}_{t - 1} &= \mathbf{f}_{t - 1} \odot \mathbf{c}_{t - 2} + \mathbf{i}_{t - 1} \odot \mathbf{g}_{t - 1}. \label{eqn:ctrelevance3}
\end{align}
(\ref{eqn:ctrelevance2}) is a one-term linearity with activation $\mathbf{o}_{t - 1} \odot \tanh(\cdot)$, while (\ref{eqn:ctrelevance3}) is a linear equation with identity activation. We compute $R(\mathbf{c}_{t - 1})$ by summing the contributions from $\mathbf{h}_t$ and $\mathbf{h}_{t - 1}$:
\[
R(\mathbf{c}_{t - 1}) = R(\mathbf{c}_t) \odot \frac{\mathbf{f}_t \odot \mathbf{c}_{t - 1}}{\mathbf{c}_t} + R(\mathbf{h}_{t - 1});
\]
and we continue the computation using (\ref{eqn:ctrelevance3}):
\[
R(\mathbf{i}_{t - 1} \odot \mathbf{g}_{t - 1}) = R(\mathbf{c}_{t - 1}) \odot \frac{\mathbf{i}_{t - 1} \odot \mathbf{g}_{t - 1}}{\mathbf{c}_{t - 1}}.
\]

LRP relevance scores have the following desirable \textit{conservation property}. Suppose $f$ is a neural network that takes inputs $x_1, x_2, \dots, x_m \in \mathbb{R}$ and produces outputs $y_1, y_2, \dots, y_n \in \mathbb{R}$. Then,
\begin{equation}
\sum_{i = 1}^n R(y_i) = \left(\sum_{i = 1}^m R(x_i) \right) + \sum_{b \in B} R(b), \label{eqn:conservation1}
\end{equation}
where $B$ is the set of bias units of $f$. This allows us to assign a scalar relevance value $r(U)$ to any collection of units $U$ by taking the sum $r(U) = \sum_{u \in U} R(u)$. For example, the collective relevance of inputs $\mathbf{x}_1$ and $\mathbf{x}_2$ is $r(\mathbf{x}_1, \mathbf{x}_2) = \mathbf{1}^\top(R(\mathbf{x}_1) + R(\mathbf{x}_2))$. In the LSTM setting, (\ref{eqn:conservation1}) then becomes
\begin{equation}
\max_i y_i = r(\mathbf{y}) = \left( \sum_{j = 1}^t r(\mathbf{x}_j) \right) + \sum_{b \in B} r(b). \label{eqn:conservation2}
\end{equation}

\section{Related Work}
\label{sec:relatedwork}

A variety of techniques exist for attribution analysis. LRP, along with \textit{DeepLIFT} \citep{shrikumarLearningImportantFeatures2017}, takes the approach of propagating a signal backwards through the network. Other methods, such as \textit{saliency analysis} \citep{simonyanDeepConvolutionalNetworks2014,liVisualizingUnderstandingNeural2016}, \textit{gradient $\odot$ input} \citep{denilExtractionSalientSentences2015}, and \textit{integrated gradients} \citep{sundararajanAxiomaticAttributionDeep2017,sundararajanAxiomaticAttributionDeep2017a}, involve computing model gradients, based on the intuition that model outputs should not be affected by changes in irrelevant features. Finally, techniques like \textit{contextual decomposition} \citep{murdochWordImportanceContextual2018} and \textit{LIMSSE} \citep{poernerEvaluatingNeuralNetwork2018} involve computing local linear approximations of the model or certain parts thereof. 

\citet{arrasEvaluatingRecurrentNeural2019} have argued, on the basis of toy tasks, that LRP yields more intuitive explanations for NLP than other techniques. We choose to use LRP for two reasons. Firstly, relevance scores may be positive or negative, allowing us to distinguish features that contribute to a model decision from those that inhibit against it. Secondly, the additive property of relevance scores allows us to aggregate relevance scores across inputs of varying lengths,  enabling us to qualitatively compare model computations on different kinds of inputs without resorting to inspection of cherry-picked examples. 

Attribution analysis is traditionally applied to sentiment analysis (e.g., \citealp{liVisualizingUnderstandingNeural2016,murdochWordImportanceContextual2018,arrasExplainingRecurrentNeural2017}), where the intrinsic sentiment value of input words gives attribution scores a natural interpretation. \citet{poernerEvaluatingNeuralNetwork2018} apply a number of attribution methods to \citeauthor{linzenAssessingAbilityLSTMs2016}'s (\citeyear{linzenAssessingAbilityLSTMs2016}) subject--verb agreement task. Whereas we use attribution analysis to assess whether models behave in an interpretable way, \citeauthor{poernerEvaluatingNeuralNetwork2018} assume that models behave in an interpretable way and evaluate attribution methods based on their ability to reveal this interpretable behavior. Like \citet{arrasEvaluatingRecurrentNeural2019}, they conclude that LRP delivers more interpretable explanations than other methods.

\section{Experimental Procedure}
\label{sec:procedure}

\begin{table*}
	\centering
	\small
	\begin{tabular}{l l l}
		\hline
		\textbf{Name} & \textbf{Template} & \textbf{Example} \\\hline
		Simple  & Det1 N1 & The \textbf{senators} \textit{laugh} \\
		Inside an Object Relative Clause (IORC) & Det2 N2 (Comp) Det1 N1 & The manager (that) the \textbf{skater} \textit{admires} \\
		Sentential Complement (SC) & Det2 N2 V Det1 N1 & The mechanic said the \textbf{manager} \textit{laughs} \\
		Across a PP (PP) & Det1 N1 P Det2 N2 & The \textbf{surgeon} in front of the ministers \textit{laughs} \\
		Across a Subject Relative Clause (SRC) & Det1 N1 Comp V Det2 N2 & The \textbf{customer} that hates the dancer \textit{laughs} \\
		Across an Object Relative Clause (ORC) & Det1 N1 (Comp) Det2 N2 V & The \textbf{officers} (that) the parents hate \textit{laugh}\\
		Short VP Coordination (SVP) & Det1 N1 V Conj & The \textbf{farmers} smile and \textit{swim} \\
		Long VP Coordination (LVP) & Det1 N1 V CompVP Conj & The \textbf{senator} knows many different foreign \\
		& & languages and \textit{is} \\\hline
	\end{tabular}
	\caption{TSE templates for subject--verb agreement.}
	\label{table:tsetemplates}
\end{table*}

Our experimental approach combines attribution analysis with the \textit{Targeted Syntactic Evaluation} (TSE) paradigm of \citet{marvinTargetedSyntacticEvaluation2018}. In TSE, models are evaluated using a testing set partitioned into subsets consisting of inputs generated from rigid syntactic templates. Model performance is then compared across the templates, revealing challenging inputs for the model. In this study, we rely on the structural rigidity imposed by the templates to aggregate LRP computations over collections of inputs.

All evaluations in this study are computed using the best-performing English language model from \citet{gulordavaColorlessGreenRecurrent2018}, which is available for download from the authors' website.\footnote{\url{https://github.com/facebookresearch/colorlessgreenRNNs}} Previous work applying TSE to LSTMs, including \citet{marvinTargetedSyntacticEvaluation2018}, \citet{shenOrderedNeuronsIntegrating2019}, and \citet{kuncoroScalableSyntaxAwareLanguage2019}, use models trained using the same data, vocabulary, and hyperparameters as \citeauthor{gulordavaColorlessGreenRecurrent2018}\footnote{The model is a 2-layer LSTM language model with 650 hidden units, 650 embedding features, and 50,000 vocabulary items, trained on a subset of English Wikipedia using stochastic gradient descent with batch size 128, dropout 0.2, and a learning rate of 20.0 with an annealing schedule based on a development set.} The remainder of this section reviews the TSE paradigm (Subsection \ref{sec:tse}) and introduces our LRP evaluation scheme (Subsections \ref{sec:tselrp} and \ref{sec:pointinggame}).

\subsection{Targeted Syntactic Evaluation}
\label{sec:tse}

TSE attempts to probe a model's syntactic knowledge through a series of diagnostic tasks. We focus here on subject--verb agreement. Test examples for subject--verb agreement are given by word sequences like (\ref{ex:agreement}).
\setcounter{exx}{\theequation}
\begin{exe}
	\ex The \textbf{keys} on the table *\textit{is}/\textit{are} \label{ex:agreement}
\end{exe}
(\ref{ex:agreement}) is a sentence truncated at a verb. In this example, the verb agrees with the noun \textit{keys}, and therefore must take the plural form \textit{are} and not the singular form \textit{is}. In TSE, the words preceding the verb, known as the \textit{preamble}, are given to the language model as input, and we compare the probability scores assigned to the singular and plural forms of the verb. If the appropriate form receives a higher score, we consider the model to have correctly predicted the number of the verb.

The templates for subject--verb agreement are shown in \autoref{table:tsetemplates}. We describe each template as a sequence of part-of-speech (POS) tags: \textit{Comp}lementizer, \textit{Conj}unction, \textit{Det}erminer, \textit{N}oun, \textit{P}reposition, \textit{V}erb, or \textit{Comp}lement of \textit{V}erb \textit{P}hrase. We label the subject of each preamble that triggers agreement on the target verb and its associated determiner with the tags N1 and Det1, respectively, and we label all other nouns and determiners with N2 and Det2, respectively. The complementizer \textit{that} appearing in the ORC and IORC templates is optional.

We make two changes to \citeauthor{marvinTargetedSyntacticEvaluation2018}'s (\citeyear{marvinTargetedSyntacticEvaluation2018}) original testing set for agreement. Firstly, the original testing set contains distinct sentences that are identical up to the preamble and the target verb, resulting in duplicate test cases. We remove those duplicate cases from our testing set. Secondly, the original testing set is completely in lowercase, even though the vocabulary is case-sensitive. We modify our testing set by capitalizing the first letter of each sentence. We will see in Section \ref{sec:results} that capitalization substantially increases our model's prediction accuracy for most templates. Although these changes render our prediction accuracies incomparable to previously reported results,\footnote{\citet{shenOrderedNeuronsIntegrating2019} also report using a slightly different testing set from \citet{marvinTargetedSyntacticEvaluation2018}.} we expect that the improvement in performance will lead to more interpretable explanations from the LRP analysis.

\subsection{LRP Evaluation}
\label{sec:tselrp}

In Section \ref{sec:lrp}, we considered LSTM classifiers, and initialized $R(\mathbf{y})$ to $\argmax(\mathbf{y})$, so that $\mathbf{1}^\top R(\mathbf{y})$ is the highest logit score produced by the model. In the current task, we are interested in \textit{comparing} the logit score assigned to one possible next word with the score assigned to another possible word. Therefore, we initialize $R(\mathbf{y})$ so that $\mathbf{1}^\top R(\mathbf{y})$ is the difference $\Delta y$ between the two scores in question. For example, when evaluating the test case given by (\ref{ex:agreement}), $R(\mathbf{y})$ is initialized as follows:
\[
R(\mathbf{y})_i = \begin{cases}
y_i, & i = \text{are} \\
-y_i, & i = \text{is} \\
0, & \text{otherwise.}
\end{cases}
\]
Observe that $\mathbf{1}^\top R(\mathbf{y}) = y_{\text{are}} - y_{\text{is}} = \Delta y$.

For each POS tag in each template, we compute the collective scalar relevance score of all words subsumed under that tag. For example, given the preamble \textit{The surgeon in front of the ministers}, we assign the P tag the relevance score $r(\text{P}) = r(\text{in}, \text{front}, \text{of})$. For an input $\mathbf{x}$, $|r(\mathbf{x})|$ represents the magnitude of $\mathbf{x}$'s contribution to $\Delta y$, and the sign of $r(\mathbf{x})$ indicates whether $\mathbf{x}$ increases $\Delta y$ or decreases it.

\subsection{Pointing Game Accuracy}
\label{sec:pointinggame}

Intuitively, we would expect that an interpretable model should assign high-magnitude attribution scores to N1 and scores close to 0 for words unrelated to subject--verb agreement. Based on this idea, \citet{poernerEvaluatingNeuralNetwork2018} propose the \textit{pointing game accuracy} as a way to induce a quantitative measure of interpretability from an attribution method. The pointing game accuracy of a model on a testing set for the agreement task is the percentage of test cases for which N1 receives the highest attribution score. Here, we compute pointing game accuracy on TSE templates using the absolute value of relevance scores.

\section{Results}
\label{sec:results}

\begin{table*}
	\small
	\centering
	\begin{tabular}{l c c c c c c c c c c}
		\hline
		\multirow{2}{*}{\textbf{Prediction Accuracy}} &  \multirow{2}{*}{\textbf{Simple}} & \textbf{IORC}  & \multirow{2}{*}{\textbf{IORC}} & \multirow{2}{*}{\textbf{SC}} & \multirow{2}{*}{\textbf{PP}} & \multirow{2}{*}{\textbf{SRC}} & \textbf{ORC}  & \multirow{2}{*}{\textbf{ORC}} & \multirow{2}{*}{\textbf{SVP}} & \multirow{2}{*}{\textbf{LVP}} \\
		& & (No \textit{That}) & & & & & (No \textit{That}) & & & \\ \hline
		\citet{marvinTargetedSyntacticEvaluation2018} & 94 & 71 & 84 & 99 & 57 & 56 & 52 & 50 & 90 & 61\\
		\citet{shenOrderedNeuronsIntegrating2019} & \textbf{100} & 81 & 88 & 98 & 68 & 60 & 51 & 52 & 92 & 74\\
		\citet{kuncoroScalableSyntaxAwareLanguage2019} & \textbf{100} & \textbf{86} & \textbf{90} & 97 & \textbf{89} & \textbf{87} & \textbf{70} & \textbf{77} & 96 & \textbf{82} \\
		Our Model (Capitalized) & \textbf{100} & 85 & \textbf{90} & 96 & 84 & \textbf{87} & 57 & 69 & \textbf{99} & 81\\
		Our Model (Lowercase) & \textbf{100} & 85 & \textbf{90} & \textbf{100} & 65 & 68 & 52 & 59 & 94 & 80 \\\hline
		\multirow{2}{*}{\textbf{Pointing Game}} &  \multirow{2}{*}{\textbf{Simple}} & \textbf{IORC}  & \multirow{2}{*}{\textbf{IORC}} & \multirow{2}{*}{\textbf{SC}} & \multirow{2}{*}{\textbf{PP}} & \multirow{2}{*}{\textbf{SRC}} & \textbf{ORC}  & \multirow{2}{*}{\textbf{ORC}} & \multirow{2}{*}{\textbf{SVP}} & \multirow{2}{*}{\textbf{LVP}} \\
		& & (No \textit{That}) & & & & & (No \textit{That}) & & & \\\hline
		Pointing Game Accuracy & 65 & 56 & 54 & 55 & 32 & 46 & {20} & {16} & {25} & {23} \\
		N2 & -- & 15 & 23 & 11 & 28 & 21 & 29 & 22 & -- & -- \\\hline
	\end{tabular}
	\caption{\textbf{Top:} Number prediction accuracies attained by our model, compared with previously reported results for similar models. \textbf{Bottom:} Pointing game accuracies attained by our model, along with the percentage of examples in which N2 received the highest-magnitude relevance score.}
	\label{table:tseaccuracy}
\end{table*}

\begin{figure*}
	\centering
	\includegraphics[trim=15 10 15 10, scale=0.42]{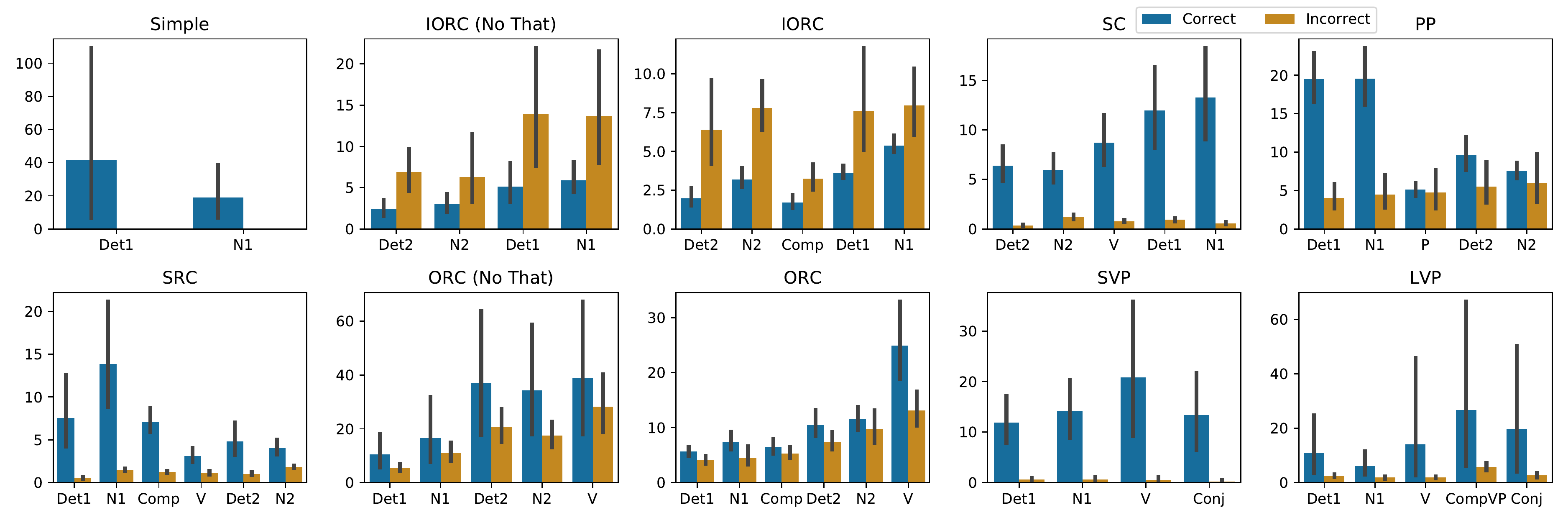}
	\caption{The absolute values of relevance scores assigned to template positions by our language model for correct (blue, left) and incorrect (orange, right) predictions.}
	\label{fig:results}
\end{figure*}

The upper portion of \autoref{table:tseaccuracy} presents our replication of \citeauthor{marvinTargetedSyntacticEvaluation2018}'s (\citeyear{marvinTargetedSyntacticEvaluation2018}) results. As mentioned in Subsection \ref{sec:tse}, capitalization improves the performance of our model on all templates except for SC and IORC. We attribute this to the fact that \textit{The} appears almost exclusively sentence-initially while \textit{the} almost never appears sentence-initially; we hypothesize that this difference in distribution provides heuristic information about which nouns are likely to be subjects. While our results are not directly comparable with previous ones, note that our model performs similarly to \citet{kuncoroScalableSyntaxAwareLanguage2019} on the capitalized inputs and \citet{shenOrderedNeuronsIntegrating2019} on the lowercase inputs.

\autoref{fig:results} shows the absolute-value relevance scores assigned to POS tags for each template. Inputs for which the model makes a correct number prediction are plotted separately from those for which the model makes an incorrect prediction. The following subsections discuss the ability of our model to identify subjects (\ref{sec:identifyingsubjects}), the role of determiners in number prediction (\ref{sec:thethatand}), the effect of polysemy on model behavior (\ref{sec:polysemy}), and an alternate strategy for agreement using verb coordination (\ref{sec:coordination}).

\begin{figure*}
	\centering
	\includegraphics[trim=15 10 15 10, scale=0.42]{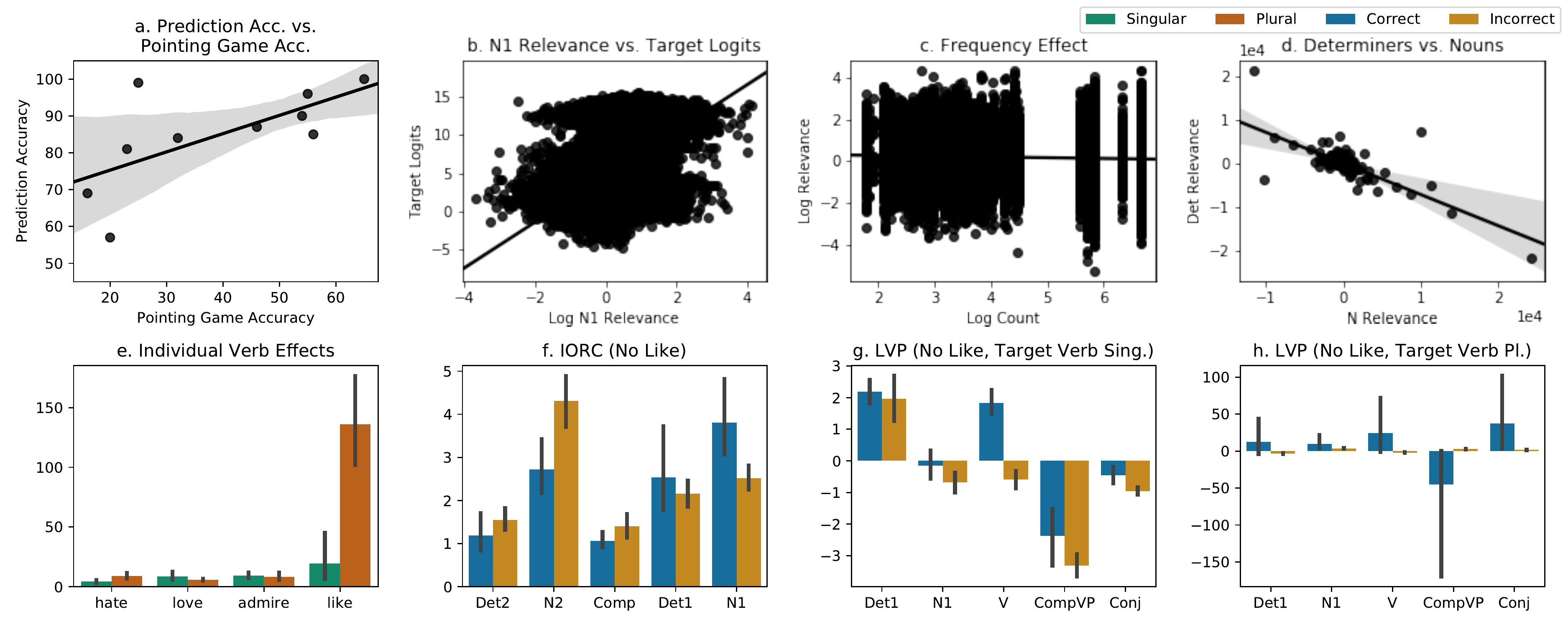}
	\caption{\textbf{a:} The relationship between pointing game accuracy and prediction accuracy ($\rho=\text{0.65}$). \textbf{b:} The relationship between N1 relevance and the logit score of the correct verb form ($\rho=\text{0.46}$). \textbf{c:} The effect of frequency on relevance ($\rho=-\text{0.08}$). \textbf{d:} The relationship between N relevance and Det relevance ($\rho=-\text{0.73}$). \textbf{e:} The absolute values of relevance scores assigned to individual verbs in ORC, plotted by their number. \textbf{f:} The absolute values of relevance scores for IORC inputs, excluding examples where the target verb is \textit{like(s)}. \textbf{g and h:} Signed relevance scores for LVP sentences without \textit{like(s)}, plotted according to the number of the target verb.}
	\label{fig:results2}
\end{figure*}

\subsection{Identifying Subjects}
\label{sec:identifyingsubjects}

The goal of this subsection is to determine whether our model makes number predictions by identifying the subject of its input, or whether it exhibits Clever Hans behavior. \autoref{fig:results} shows that in most cases, $|r(\text{N1})| > |r(\text{N2})|$ when the model makes a correct number prediction, and \textit{vice versa}. When the model makes incorrect predictions, $|r(\text{N2})|$ is closer to $|r(\text{N1})|$, indicating that these are situations in which the model does not confidently distinguish one noun from the other.

There are two exceptions to this pattern. In IORC, we have $|r(\text{N1})| > |r(\text{N2})|$ even when the network makes incorrect predictions, and in ORC, we have $|r(\text{N2})| > |r(\text{N1})|$ even when the network makes correct predictions. We will see in Subsection \ref{sec:polysemy} that the IORC phenomenon is due to the polysemous nature of the target verb \textit{like(s)}, which is amply present in the test set. With ORC, we note that our model achieves the lowest performance on this template, with an average accuracy of 66\% on inputs both with and without the complementizer \textit{that}. 

In \autoref{fig:results2}a, we see that pointing game accuracy on templates is positively correlated with prediction accuracy, with a Pearson correlation coefficient of $\rho = \text{0.65}$. Similarly, the percentage of inputs matching a given template for which N2 receives the highest-magnitude relevance score is negatively correlated with prediction accuracy, with $\rho = -\text{0.65}$. This suggests that the ability to identify N1 and distinguish it from N2 is an important factor in determining the model's ability to perform number prediction. Overall, \autoref{fig:results2}b shows that $\log(|r(\text{N1})|)$ is correlated with the logit score of the correct verb form, while \autoref{fig:results2}c shows that word frequency in the training set, a possible source of spurious correlation, has no significant effect on relevance.

\subsection{Determiners}
\label{sec:thethatand}

A striking observation about \autoref{fig:results} is that $|r(\text{Det1})|$ and $|r(\text{Det2})|$ are often greater than or close to $|r(\text{N1})|$ and $|r(\text{N2})|$, respectively. This phenomenon is unexpected, since the determiner \textit{The}/\textit{the}, which is the only possible value for Det1 and Det2, does not carry agreement information.

\autoref{fig:results2}d shows that $r(\text{N})$ is negatively correlated with $r(\text{Det})$. The regression line gives us
\[
r(\text{Det}, \text{N}) = \text{0.287}r(\text{N}) + \text{0.642},
\]
which indicates that when \textit{The}/\textit{the} is combined with a noun, it asymptotically has the effect of scaling the relevance of the resulting noun phrase by a factor of roughly 0.3. The regression line predicts that $r(\text{Det}, \text{N})$ may have the opposite sign of $r(\text{N})$ when $-\text{2.235} < r(\text{N}) < \text{0}$. In this region, the negative relevance of N is overruled by the positive relevance of Det. This occurs in 22\% of cases, and we will later see that it plays an important role in the LVP template.


\subsection{Verbs and Polysemy}
\label{sec:polysemy}

\begin{table*}
	\begin{subtable}{0.5\textwidth}
		\centering
		\small
		\begin{tabular}{l c c}
			\hline
			\textbf{Template} & \textbf{All Inputs} & \textbf{No \textit{Like}} \\\hline
			IORC (No \textit{That}) & 85 & \textbf{93} \\
			IORC & 90 & \textbf{94} \\
			SRC & 87 & \textbf{88} \\
			ORC (No \textit{That}) & 57 & \textbf{58} \\
			ORC & 69 & \textbf{71} \\
			LVP & 81 & \textbf{95} \\\hline
		\end{tabular}
		\caption{Prediction accuracies with inputs containing \textit{like(s)} removed. Simple, SC, PP, and SVP do not have any inputs with \textit{like(s)}.}
		\label{table:nolike}
	\end{subtable}
	\begin{subtable}{0.5\textwidth}
		\centering
		\small
		\begin{tabular}{l l}
			\hline
			\textbf{V} & \textbf{CompVP} \\\hline
			know(s) & many different foreign languages \\
			like(s) & to watch television shows \\
			is/are & twenty three years old \\
			enjoy(s) & playing tennis with colleagues \\
			write(s)  & in a journal every day\\\hline
		\end{tabular}
		\caption{Values for V and CompVP in LVP inputs.}
		\label{table:vpinputs}
	\end{subtable}
	\caption{}
\end{table*}

\begin{table}
	\centering
	\small
	\begin{tabular}{l c c c}
		\hline
		\textbf{Template } & \textbf{Likes} & \textbf{Like} & \textbf{Other} \\\hline
		IORC (No \textit{That}) & 0 & \textbf{63} & 37 \\
		IORC  & 0 & \textbf{55} & 45 \\
		LVP & 9 & 29 & \textbf{62} \\\hline
	\end{tabular}
	\caption{The percentage of incorrect predictions where the predicted verb is \textit{likes}, \textit{like}, or some other verb.}
	\label{table:incorrectlike}
\end{table}

On average, ORC predictions assign the highest-magnitude relevance scores to V. In correct predictions without \textit{that}, this relevance magnitude is disproportionately high, with $|r(\text{V})| \approx \text{2} \cdot |r(\text{N2})|$. Considering the relevance magnitudes assigned to individual verb types, \autoref{fig:results2}e reveals that the plural form \textit{like} receives far more relevance than other verbs. Intuitively, the fact that \textit{like} can be used as a verb, preposition, noun, or adjective means that this particular form conveys information to the model that is not conveyed by the other verb forms, affecting its behavior.

\autoref{table:nolike} shows that for all templates using the verb \textit{like(s)}, model performance improves when examples containing \textit{like(s)} are omitted. SRC and ORC experience modest improvements, while IORC and LVP improve substantially. The former two contain \textit{like(s)} in the preamble, but not in the target verb; IORC contains \textit{like(s)} in the target verb, but not in the preamble; and LVP contains \textit{like(s)} both in the target verb and in the preamble. This indicates that including \textit{like(s)} as the target verb is a major source of errors in number prediction. Intuitively, the polysemy of \textit{like} means that it may appear in more contexts than its singular counterpart \textit{likes}. For example, \textit{like a brother\dots} is a reasonable continuation of \textit{The customer that loves the dancer} in which \textit{like} is used as a preposition. The existence of these alternatives may therefore bias the model in favor of predicting \textit{like} over \textit{likes}. This is indeed borne out in \autoref{table:incorrectlike}: the model is far more likely to incorrectly predict \textit{like} than \textit{likes} for LVP, and the model never incorrectly predicts \textit{likes} for IORC. Whereas \autoref{fig:results} showed that IORC anomalously exhibits $|r(\text{N1})| > |r(\text{N2})|$ when making incorrect predictions, in \autoref{fig:results2}f we see that this behavior is entirely due to the bias toward \textit{like}.

\subsection{Coordination}
\label{sec:coordination}

Two templates with prominently low pointing game accuracies are SVP and LVP. Here, V and the conjunction \textit{and} receive high-magnitude relevance scores. We interpret the model to be relying on the fact that coordinated verbs generally agree with the same subject, and therefore bear the same number agreement morphology. Thus, these templates present an example in which the model employs a robust alternative strategy that does not necessarily require identifying N1, but instead obtains number features from the verb. 

Figures \ref{fig:results2}g and \ref{fig:results2}h present a more detailed analysis of LVP, excluding examples involving \textit{like(s)}. In most cases, CompVP receives a negative relevance score, which is offset by $r(\text{Det1}, \text{N1}, \text{V})$. As shown in \autoref{table:vpinputs}, CompVP always contains at least one noun, which may be either singular or plural, thus distracting the model with confounding agreement information. The model makes incorrect predictions when $r(\text{V})$ is too close to 0 to offset $r(\text{CompVP})$. Observe that $r(\text{V})$ by itself is not large enough to completely offset $r(\text{CompVP})$: it is only able to do so by combining with $r(\text{Det1},\text{N1})$. Thus, while N1 does not receive the highest-magnitude relevance scores, it is still important for ensuring correct model behavior. When the target verb is singular, $r(\text{N1})$ is slightly negative; but because its magnitude is small enough to lie within the $-\text{2.235} < r(\text{N1}) < \text{0}$ zone, $r(\text{Det1})$ reverses its directionality. In this situation, we may understand Det1 to provide a correction for the case where $r(\text{N1})$ is negative but close to 0. Finally, in addition to Det1, N1, V, and CompVP, the conjunction \textit{and} often receives a high-magnitude relevance score. In \autoref{fig:results2}h we see that this score is only positive when the target verb is plural. Thus, we may view \textit{and} as providing a plural bias to the network, possibly arising from the inherent plurality of coordinated subjects.

\section{Discussion}
\label{sec:discussion}

The analysis we have presented reveals several key insights about LSTM behavior. Owing to the additive nature of both LRP and the cell state update equation, we may view LSTMs as devices that accumulate information extracted from inputs. Our model's ability to make correct predictions is determined by its ability to weigh information about N1 with information about N2. The IORC, SC, PP, and SRC templates, which consist of two noun phrases with some material in between, show that the model is able to adjust the relative prominence of the two noun phrases to fit the syntactic context. This behavior is also seen with determiners, which serve to adjust the magnitude and sometimes the direction of the relevance introduced by nouns. Finally, as we have seen with LVP, the model is able to combine information from multiple sources to justify one decision over another.

Our observation that $|r(\text{N1})|$ is often close to $|r(\text{N2})|$ when the model makes incorrect predictions is consistent with two findings of \citet{giulianelliHoodUsingDiagnostic2018} regarding the encoding of agreement information in the hidden state vector $\mathbf{h}_t$. Firstly, \citeauthor{giulianelliHoodUsingDiagnostic2018} claim that agreement errors are often due to misencodings of the subject. This is reflected in our framework by the fact that incorrect predictions often result in a smaller N1 relevance (e.g., in the PP template), indicating that the model has failed at least in part to extract number information from N1. Secondly, \citeauthor{giulianelliHoodUsingDiagnostic2018} observe that when making incorrect predictions, $\mathbf{h}_t$ loses its number encoding after the second noun has been processed. This may be explained by the additive nature of relevance: if $|r(\text{N1})| \approx |r(\text{N2})|$ in a sentence where N1 and N2 bear opposite number features, then we expect that $r(\text{N1}) + r(\text{N2}) \approx \text{0}$. Thus, the agreement information extracted from N1 neutralizes the information extracted from N2.

No explanation has been offered in Section \ref{sec:results} for the relevance scores of the ORC template. Given that both V and N2 receive high-magnitude relevance scores even though they transmit the same agreement information, model behavior on this template is likely determined entirely by N2. Thus, we conclude that our model cannot handle the ORC construction, and that its slight improvement over chance is due to Clever Hans behavior.

\section{Conclusion}
\label{sec:conclusion}

Our analysis has shown that the LSTM language model of \citet{gulordavaColorlessGreenRecurrent2018} enforces subject--verb agreement in an interpretable manner. While the model draws number information from several sources, including nouns, verbs, and \textit{and}, identifying the target verb's subject is crucial to the model's ability to execute the agreement task, even in cases where it is used in conjunction with evidence from other inputs. In the case where the model is unable to identify the subject, namely ORC, the model only slightly outperforms chance. These findings demonstrate that the successes of the LSTM language model on the agreement task are \textit{not} due to Clever Hans behavior.

The methodological approach we have taken in this study is based on the synthesis of two existing methodologies: experimentally controlled testing sets and attribution analysis. Both components are required for our approach. For instance, while \citet{poernerEvaluatingNeuralNetwork2018} report a pointing game accuracy of 86\% in their number prediction study,\footnote{Their result is not comparable with ours, since their study treats number prediction as a supervised binary classification task rather than an evaluation scheme for language models.} it is difficult to discern the significance of this number on its own without further context. By correlating pointing game accuracy with prediction accuracy on different kinds of testing sets, however, we can determine the extent to which model behavior results from executing the desired strategy. Thus, combining multiple analytical techniques may prove to be a fruitful way to gain insights beyond what is revealed by standard evaluation metrics.

Although we have argued that positive results about the grammatical abilities of LSTMs are not due to Clever Hans behavior, we have not shown that LSTMs are able to infer stack-like representations of hierarchical syntactic structure or memory-bounded approximations thereof. While TSE introduces syntactic complexity by incorporating a diversity of constructions, the TSE templates are generally quite simple in terms of embedding depth and dependency length. As discussed in Section \ref{sec:discussion}, most of the templates are superficially similar to one another even if they are represented differently in linguistic theory. Among the rest, LVP features a long-distance dependency between the target verb and N1, while ORC features a center-embedding construction. The model's promising performance on LVP and failure on ORC seems to suggest that depth is more challenging for the model than distance. Such issues provide avenues of exploration for future work.

\bibliography{bibliography}
\bibliographystyle{acl_natbib}

\appendix

\end{document}